%% file: arXiv.tex
\documentclass[11pt]{article}
\usepackage{jmlr2e-mod}
\pdfoutput=1

\usepackage{graphicx}
\usepackage{booktabs}
\usepackage{subcaption}

\title{Using Neural Architecture Search for Improving Software Flaw Detection in Multimodal Deep Learning Models}

\author{\name Alexis Cooper \email acoope@sandia.gov\\
	\name Xin Zhou \email xzhou1@sandia.gov\\
	\name Scott Heidbrink \email sheidbr@sandia.gov\\
	\name Daniel M. Dunlavy\footnotemark[2] \email dmdunla@sandia.gov\\
	\addr Sandia National Laboratories\\
	\addr Albuquerque, NM 87123, USA}

\begin{document}
		
	\maketitle

	\renewcommand*{\thefootnote}{(\fnsymbol{footnote})}
	\footnotetext[2]{Corresponding author.}
	\renewcommand*{\thefootnote}{\arabic{footnote}.}
	
\begin{abstract}
\input{abstract}
\end{abstract}


\begin{figure}[b!]
\centering
\includegraphics[width=\textwidth]{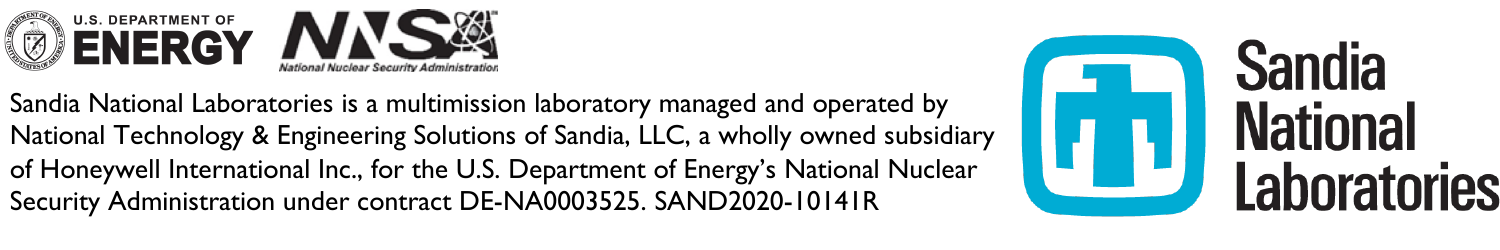}
\end{figure}

\section{Introduction}
\label{sec:intro}
\input{intro}

\section{Methods}
\label{sec:methods}
\input{methods}

\section{Experiments}
\label{sec:exp}
\input{exp}

\section{Results}
\label{sec:results}
\input{results}

\section{Conclusions}
\label{sec:conc}
\input{conc}



\clearpage
\appendix
\section{NAS-GDAS-JAE Cell Search Results}
\label{sec:app}
\input{app_a}

\end{document}

%% file: abstract.tex
Software flaw detection using multimodal deep learning models has been demonstrated as a very competitive approach on benchmark problems. In this work, we demonstrate that even better performance can be achieved using neural architecture search (NAS) combined with multimodal learning models. We adapt a NAS framework aimed at investigating image classification to the problem of software flaw detection and demonstrate improved results on the Juliet Test Suite, a popular benchmarking data set for measuring performance of machine learning models in this problem domain.

%% file: intro.tex
Most current approaches for software flaw detection rely on analysis of a single representation of a software program (e.g., source code or program binary compiled in a specific way for a specific hardware architecture).  Recent work using multiple software representations and multimodal deep learning illustrates the benefits of leveraging both source and binary information in detecting flaws~\cite{HeRoDu20}. However, when using deep learning models, determining the most effective neural network architecture can be a challenge. Neural architecture search (NAS) is one way to perform an automated search across many different neural network architectures to find improved model architectures over manually-designed ones.  In this work, we use a gradient-based NAS method that leverages a differentiable architecture sampler (GDAS)~\cite{DoYa2019}, which was identified as the best NAS method across 10 popular approaches when applied to image classification problems~\cite{DoYa2020}. 

The remainder of this report is organized as follows. In Section~\ref{sec:methods}, we provide an overview of the multimodal deep learning and NAS methods used to create flaw detection models. In Section~\ref{sec:exp}, we define the set of experiments conducted to assess performance of these models over the baseline of not using NAS. In Section~\ref{sec:results}, we present the results of these experiments on a standard benchmark data set used in flaw detection research. And, finally, in Section~\ref{sec:conc}, we summarize our conclusions and provide suggestions for future work in this area.

%% file: methods.tex
In this section, we describe the Joint Autoencoder (JAE) multimodal deep learning model for software flaw detection~\cite{HeRoDu20} and the cell-based neural architecture search (NAS) approach used to determine an optimal architecture for that model.

\subsection{Multimodal Deep Learning for Software Flaw Detection}
\begin{figure}[!ht]
	\centering
	\includegraphics[width=\linewidth]{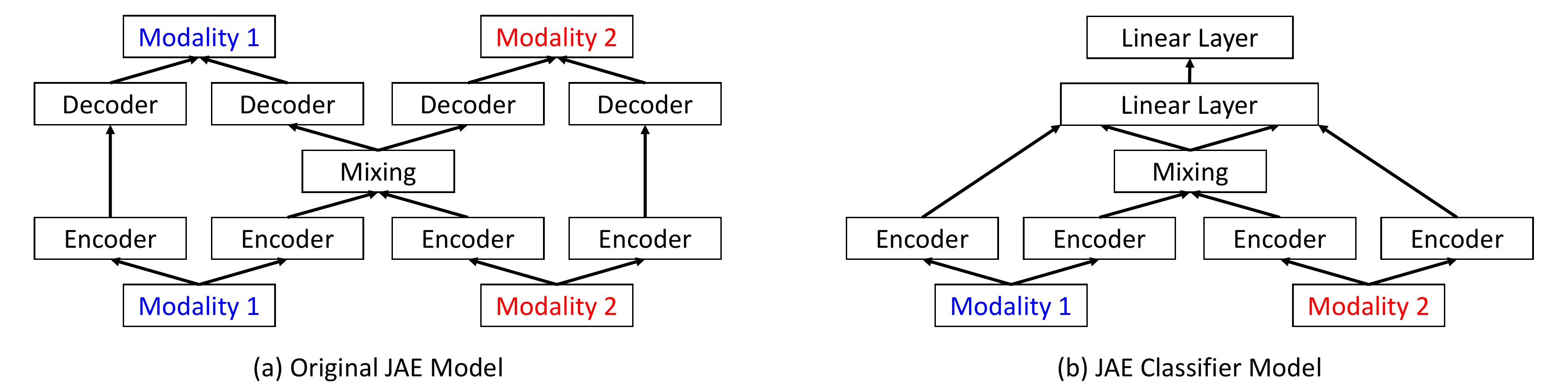}
	\caption{JAE Structure}
	\label{fig:JAE_arch}
\end{figure}

The neural network architecture selected for these experiments is an early fusion multimodal learning model called Joint Autoencoder (JAE) \cite{epstein2018joint}. JAE was originally developed for learning multiple tasks simultaneously based on sharing features that are common to all tasks. 
Figure~\ref{fig:JAE_arch}(a) illustrates the architecture of the original JAE model, which contains 2 encoder/decoder components per modality and a single mixing component that combines the output from one of the encoders associated with each modality.  The components that do not interact with the mixing component are referred to as {\em private branches}~\cite{epstein2018joint}. Note that each of the components depicted in the image (i.e., each box in the image) can contain one or more traditional neural network layers. Recently, an adaptation of the JAE model, referred to here as the JAE Classifier Model, was developed for classifying software functions as to whether or not they contain flaws/bugs~\cite{HeRoDu20}. 
Figure~\ref{fig:JAE_arch}(b) illustrates the architecture of the JAE Classifier Model, where we remove the decoders and use a linear layer to concatenate the outputs from previous layers. In the JAE Classifier Model, we use one or more linear layers with LeakyReLU activation for encoders and the mixing components. In the first linear layer, the number of input features will be the total length of two private branch encoders plus the number of output features from mixing component, and the number of output features is fixed as 50. In the final linear layer, a classifier layer is used, mapping 50 input features to the number of classes. In the flaw detection models used here, we use two classes, {\em flawed} and {\em not flawed}.

\subsection{Neural Architecture Search}
 The JAE architectures described in the previous section were designed manually and thus may not be optimal for the learning tasks to which they are applied. To address this potential issue, we leverage a Neural Architecture Search (NAS) strategy to determine an optimal architecture for the flaw detection task. The specific form of NAS we employ here is based on cell-based search, in which a cell represents a portion of the architecture and is defined using a densely-connected directed acyclic graph (DAG)~\cite{DoYa2020}. The edges of the DAG represent architecture layers and the nodes represent sums of the feature maps output from each of those layers. The search is performed over a set of operations (i.e., network layers) and the weights associated with those operations. Optimization of the cell structure and weights is performed within each iteration of the overall model training.
 
\begin{figure}[!ht]
	\centering
	\includegraphics[width=\linewidth]{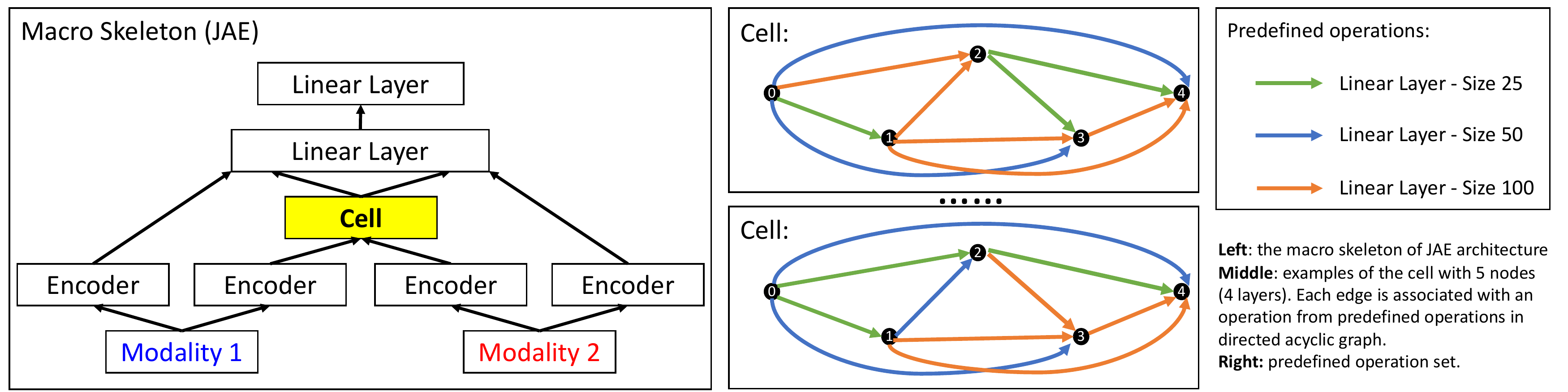}
	\caption{JAE Structure}
	\label{fig:JAE_structure}
\end{figure}

 In this work, we define the macro skeleton, i.e., the full NAS architecture, as the JAE Classifier Model and the cell as the mixing layer with that model. Figure~\ref{fig:JAE_structure} illustrates the macro skeleton architecture (left), example DAG instances of the cell (center), and the cell operations used in our work (right). As noted in the image, the cell operations consist of single linear layers of sizes 25, 50, and 100 (i.e., the number of nodes in the layer). Details of the interpretation of the cell examples as sums of the feature maps of the operations can be found in~\cite{DoYa2019}.
 
 We adapt the Automated Deep Learning (AutoDL) NAS comparison  framework\footnote{https://github.com/D-X-Y/AutoDL-Projects}, which implements the NAS-BENCH-201 \cite{DoYa2020} image classification benchmark, for use with our flaw detection classification problem. As recommended in the NAS-BENCH-201 experiments on images and confirmed in preliminary experiments with the JAE Classifier Model, we use the GDAS search strategy~\cite{DoYa2019} in the work presented here. GDAS is a gradient-based search method using differentiable architecture sampler to optimize the cell search, and it has been demonstrated to be one of the more efficient NAS techniques that relies on more than simple random sampling for the cell search. 
 
Optimization of the weights in the cell layers is performed using stochastic gradient descent (SGD)~\cite{sgd} and the overall macro skeleton architecture model fitting is performed using the ADAM optimizer~\cite{adam}, both as implemented in the AutoDL framework.

%% file: exp.tex
In this section, we describe the experiments we performed to answer the following questions: 

\begin{itemize}
	\item Are there differences between handcrafted JAE structure and selected structure from NAS?
	\item Are there improvements on flaw  detection performance after implementing NAS?
\end{itemize}

\subsection{Data}

As we are measuring potential improvements when using NAS on the JAE Classifier Model, we use the same subset of the Juliet Test Suite data~\cite{NIST} from the software flaw detection experiments performed in~\cite{HeRoDu20}. The Juliet Test Suite \cite{NIST} is a collection of test cases in the C/C++ language, providing pairs of functions with and without software flaws. The test cases laws are organized into collections based on the Common Weakness Enumeration (CWEs) of the specific flaws exhibited in each function. Table~\ref{tab:juliet} lists the test case CWE collections used in this work. This set of test cases represents a wide range of the types of flaws found in real-world software systems. We use the features extracted from this data as defined in~\cite{HeRoDu20}.

In our experiments, we split each CWE collection into three data sets: 80\% train, 10\% validation, and 10\% test. For cell search, we use the train and validation data sets to search for the best cell.

\begin{table}[ht!]
	\centering
	
	\begin{tabular}{|c|l|l|l|}
		\hline
		\textbf{CWE} & \textbf{Flaw Description} & \textbf{\# Flawed} & \textbf{\# Not Flawed}\\
		\hline
		121 & Stack Based Buffer Overflow & 6346 & 16868 \\
		\hline
		190 & Integer Overflow                 & 3296 & 12422  \\
		\hline
		369 & Divide by Zero                    & 1100 & 4142  \\
		\hline
		377 & Insecure Temporary File      & 146 & 554  \\
		\hline
		416 & Use After Free                    & 152 & 779  \\
		\hline
		476 & NULL Pointer Dereference    & 398 & 1517  \\
		\hline
		590 & Free Memory Not on Heap   & 956 & 2450  \\
		\hline
		680 & Integer to Buffer Overflow   & 368 & 938  \\
		\hline
		789 & Uncontrolled Mem Alloc       & 612 & 2302  \\
		\hline
		78   & OS Command Injection       & 6102 & 15602  \\ 
		\hline
	\end{tabular}
	\caption{Juliet Test Suite Data Summary}
	\label{tab:juliet}
\end{table}

\subsection{Methods used in Experiments}
We compare flaw detection results using the JAE Classifier Model and application of the GDAS to the cell-based macro skeleton described in the previous section. The manually-designed JAE Classifier Model used a mixing component with a single linear layer consisting of 50 nodes, and we refer to this model as the {\em JAE-Mixing-50} model. In our experiments, we also investigated the use of a larger layer of size 100, and we refer to that model here as the {\em JAE-Mixing-100} model. The GDAS-based model is referred to here as the NAS-GDAS-JAE model. 

\subsection{Measurements used in Comparing Methods}
For each of the Juliet Test Suite CWE collections, we performed $N \times 2$ cross validation~\cite{nx2cv} with $N=5$. We use this form of cross validation as it provide a pessimistic estimate of the generalization error; when training models for operational use, we often use more than 50\% of our training data to fit the final model. We use class-averaged accuracy---the average of the accuracies of instances from each class, normalized by the size of each class---to adjust for the skew in the sizes of the {\em flawed} and {\em not flawed} instance (see Table~\ref{tab:juliet} for details). This approach addresses skew by not favoring classification results from either of the classes when they are not equal in size. For each method, we compute and report the sample mean and sample standard deviation of the class-averaged accuracy results for each method on each CWE collection.

\subsection{Cell Structure Optimization}
As mentioned earlier, in the NAS-GDAS-JAE model, cell search is performed using SGD optimization. The specific parameters used in the AutoDL implementation of SGD are provided in Table~\ref{fig:param}.

\begin{table}[!h]
	
	\centering
	\begin{tabular}{|l|l|}
		\hline
		\textbf{Parameter} & \textbf{Value} \\
		\hline
		scheduler & cos \\ 
		\hline
		LR & 0.0005 \\
		\hline
		eta\_min & 0.001 \\
		\hline
		epochs & 100 \\
		\hline
		optim & SGD \\
		\hline
		decay & 0.000001 \\
		\hline
		momentum & 0.9 \\ 
		\hline
		nesterov & 1 \\
		\hline
		criterion & Softmax \\
		\hline
		batch\_size & 32 \\
		\hline

	\end{tabular}

\caption{NAS-GDAS-JAE Parameters for SGD Cell Search}
	
\label{fig:param}
\end{table}

\subsection{Cell Structure Representation}
\begin{figure}[!h]
	\centering
	\includegraphics[width=\linewidth]{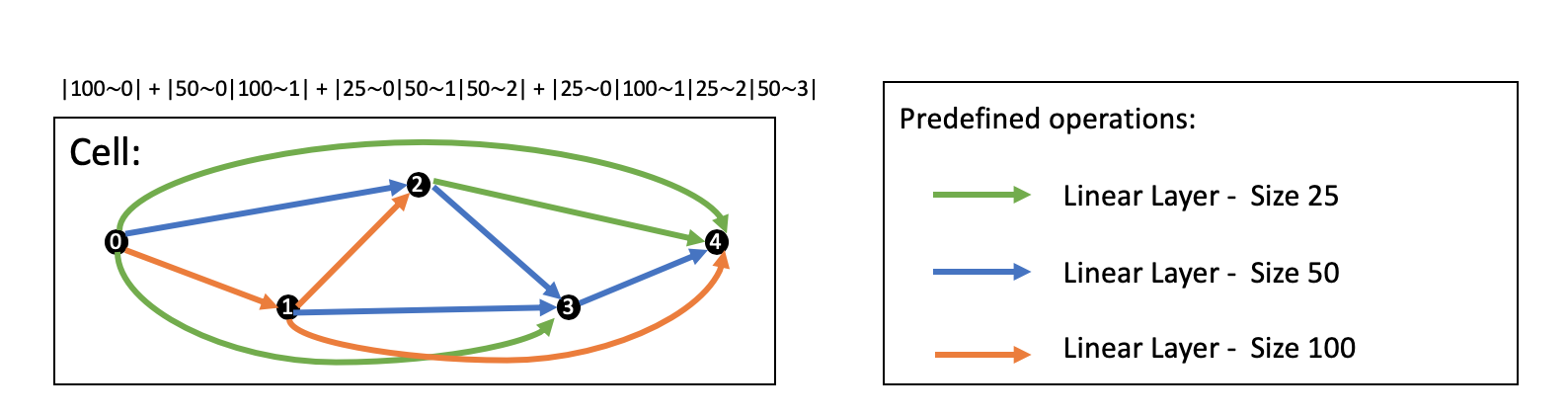}
	\caption{Example Cell Structure in the NAS-GDAS-JAE Model}
	\label{fig:sample}
\end{figure}

The result of the cell search in the NAS-GDAS-JAE model is a DAG representing several linear layers of different sizes (based on our defined cell operations). The AutoDL framework in which we implemented NAS-GDAS-JAE represents a DAG instance using a string to define the specific cell operations and sums of feature maps. Figure~\ref{fig:sample} illustrates the string output of an example DAG, which is

{\centering
{\small
\begin{verbatim}
|100~0| + |50~0|100~1| + |25~0|50~1|50~2| + |25~0|100~1|25~2|50~3|  .
\end{verbatim}
}
}
This summands in the string represent the sums of the feature maps associated with different cell operations. Each sum is defined inside the ``\verb#|#  \verb#|#'' delimiters, where each cell operation and the edge source node is listed. For example, the summand in the example above of ``{\small\verb#|25~0|50~1|50~2|#}'' represents the sum of the feature maps of three cell operations (i.e., linear layers) at node 3 as depicted in the image---the green edge (size 25) from node 0, the blue edge (size 50) from node 1, and the blue edge (size 50) from node 2. 

%% file: results.tex
In this section, we present the results of our experiments leveraging multimodal learning models and neural architecture search to address the question of software flaw detection.

\subsection{Optimized Cell Structure of NAS-GDAS-JAE Models}

The optimized cell structure of the NAS-GDAS-JAE models for each of the Juliet Test Suite data sets can be found in Table~\ref{tab:NAS-cell}. Note that none of the final cell structures across the difference data sets are the same.

\begin{table}[!h]
\caption{GDAS-JAE Search Results
}
\label{tab:NAS-cell}
\centering
\resizebox{\textwidth}{!}{	
\begin{tabular}{|c|l|}
	\hline

\textbf{CWE} &\textbf{Cell Structure (string representation of DAG)}\\
	\hline
	121 &\footnotesize{\texttt{|100$\sim$0| + | 50$\sim$0|100$\sim$1| + | 25$\sim$0| 50$\sim$1| 50$\sim$2| + | 25$\sim$0|100$\sim$1| 25$\sim$2| 50$\sim$3|}}  \\
	
	\hline
		190 &\footnotesize{\texttt{| 50$\sim$0| + |100$\sim$0| 25$\sim$1| + | 25$\sim$0| 25$\sim$1| 50$\sim$2| + |100$\sim$0| 50$\sim$1|100$\sim$2| 25$\sim$3|}} \\
		
	\hline
		369 &\footnotesize{\texttt{| 25$\sim$0| + | 25$\sim$0|100$\sim$1| + | 25$\sim$0|100$\sim$1| 25$\sim$2| + | 50$\sim$0| 25$\sim$1| 25$\sim$2|100$\sim$3|}} \\
	\hline
		377 &\footnotesize{\texttt{| 50$\sim$0| + | 25$\sim$0| 25$\sim$1| + | 50$\sim$0| 25$\sim$1|100$\sim$2| + |100$\sim$0|100$\sim$1| 50$\sim$2| 25$\sim$3|}} \\
	\hline
		416 &\footnotesize{\texttt{| 25$\sim$0| + | 50$\sim$0|100$\sim$1| + | 50$\sim$0|100$\sim$1|100$\sim$2| + | 25$\sim$0|100$\sim$1|100$\sim$2| 50$\sim$3| }}   \\
	\hline
		476 &\footnotesize{\texttt{|100$\sim$0| + |100$\sim$0| 50$\sim$1| + | 25$\sim$0| 50$\sim$1| 25$\sim$2| + | 50$\sim$0| 25$\sim$1| 50$\sim$2| 50$\sim$3|}}    \\
	\hline
		590 &\footnotesize{\texttt{|100$\sim$0| + | 50$\sim$0| 25$\sim$1| + | 50$\sim$0|100$\sim$1|100$\sim$2| + |100$\sim$0| 25$\sim$1| 50$\sim$2|100$\sim$3|}}\\
	\hline
		680 &\footnotesize{\texttt{|100$\sim$0| + |100$\sim$0| 50$\sim$1| + | 50$\sim$0|100$\sim$1|100$\sim$2| + | 50$\sim$0| 50$\sim$1| 50$\sim$2| 25$\sim$3|}}    \\
	\hline
		78 &\footnotesize{\texttt{|100$\sim$0| + | 50$\sim$0|100$\sim$1| + |100$\sim$0|100$\sim$1| 50$\sim$2| + |100$\sim$0| 50$\sim$1| 25$\sim$2| 50$\sim$3|}}   \\
	\hline
		789 &\footnotesize{\texttt{| 25$\sim$0| + | 25$\sim$0| 25$\sim$1| + | 25$\sim$0| 50$\sim$1|100$\sim$2| + | 50$\sim$0| 50$\sim$1|100$\sim$2|100$\sim$3|}} \\
	\hline
\end{tabular}
}
\end{table}

The differences in cell structures may be due to the fact that the cell search is a global optimization problem, but the SGD method is only guaranteed to find a local optimizer. Or this may be due to the differences between the data associated with the different flaw types. More work is needed to better understand the source for these differences. To illustrate some of the differences, we present plots of the convergence behaviors of the cell search (search) and macro skeleton architecture (eval) optimizations in Appendix~\ref{sec:app}. Over 100 epochs, we see a wide range of behaviors, maximum accuracy values achieved, and search/eval differences across the various data sets. More work is needed to better understand how these convergence behaviors impact the flaw detection results in general.

\subsection{Flaw Detection Results}

Table~\ref{table:comparison} shows the flaw detections results using the three models descried above. The two {\em JAE-Mixing-N} models (with $N=50$ and $N=100$) are considered baselines for the NAS-GDAS-JAE model, as they use the manually-designed architecture described in previous results~\cite{HeRoDu20}. The results listed in the table are the sample means and sample standard standard deviations of the class averaged accuracy per Juliet Test Suite data set. The boldfaced results indicate the best mean class-averaged accuracy for each data set (i.e., per row). Note that many of the differences between the means are not separated by more than a single sample standard deviation (across methods/columns), and thus the improvements using NAS may not be statistically significant. More work is need to determine if these improvements generalize and are statistically significant.

\begin{table}[h!]
	
	\caption{Sample means and standard deviations of class averaged accuracy using $5 \times 2$ cross validation (boldfaced results are best across methods for each data set)}

	\label{table:comparison}
	\centering
	\begin{tabular}{|c|c|c|c|}
		\hline

		\textbf{CWE}& \textbf{JAE-Mixing-50 } & \textbf{JAE-Mixing-100 } & \textbf{NAS-GDAS-JAE} \\
\hline
121 & 0.9972$\pm$0.0009 & \textbf{0.9975}$\pm$0.0008 & 0.9970$\pm$0.0012 \\
\hline
190 & 0.9867$\pm$0.0068 & \textbf{0.9907}$\pm$0.0059 & 0.9884$\pm$0.0067 \\
\hline
369 & 0.9485$\pm$0.0206 & 0.9500$\pm$0.0203 & \textbf{0.9703}$\pm$0.0220 \\
\hline
377 & 0.9309$\pm$0.0614 & 0.9285$\pm$0.0420 & \textbf{0.9514}$\pm$0.0410 \\
\hline
416 & 0.9074$\pm$0.0620 & 0.9359$\pm$0.0471 & \textbf{0.9468}$\pm$0.0400 \\
\hline
476 & 0.9991$\pm$0.0019 & \textbf{1.0000}$\pm$0.0000 & \textbf{1.0000}$\pm$0.0000 \\
\hline
590 & \textbf{1.0000}$\pm$0.0000 & \textbf{1.0000}$\pm$0.0000 & \textbf{1.0000}$\pm$0.0000 \\
\hline
680 & 0.9344$\pm$0.0139 & 0.9356$\pm$0.0115 & \textbf{0.9417}$\pm$0.0197 \\
\hline
78 & 0.9398$\pm$0.0110 & 0.9360$\pm$0.0143 & \textbf{0.9427}$\pm$0.0155 \\
\hline
789 & 0.9672$\pm$0.0201 & 0.9630$\pm$0.0183 & \textbf{0.9683}$\pm$0.0215 \\
\hline
		
	\end{tabular}
	
\end{table}

%% file: conc.tex
In this work, we implemented a cell-based neural architecture search strategy to improve upon a manually-designed multimodal learning model for software flaw detection. Our results indicate that NAS leads to improved multimodal models that are specific to the software data being analyzed. These preliminary results provide a starting point for leveraging NAS for such a problem, as there are many open questions that still need to be addressed. In the work presented here, we used a cell that replaces only a small part of the JAE Classifier Model from~\cite{HeRoDu20}. However, larger, more complicated cells could lead to more pronounced improvements, but this would come at increased optimization and training cost as well. Determining the trade-offs between cell complexity and computational cost could be a useful research activity.

%% file: app_a.tex
\begin{figure}[!h]
	\centering
	\begin{subfigure}[b]{0.48\linewidth}
		\includegraphics[width=\linewidth]{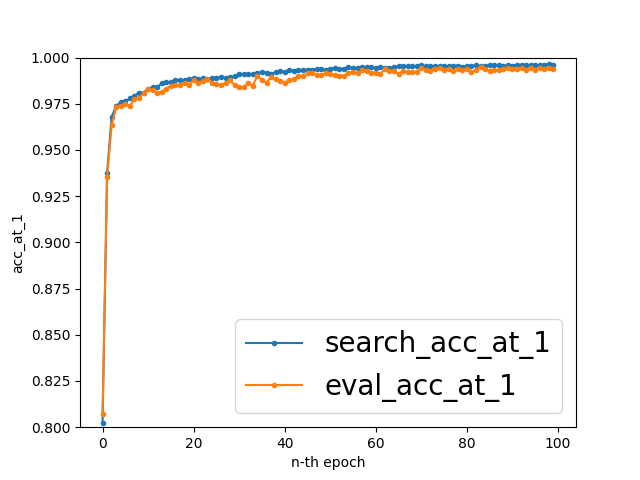}
		\caption{CWE-121}
	\end{subfigure}
	\begin{subfigure}[b]{0.48\linewidth}
		\includegraphics[width=\linewidth]{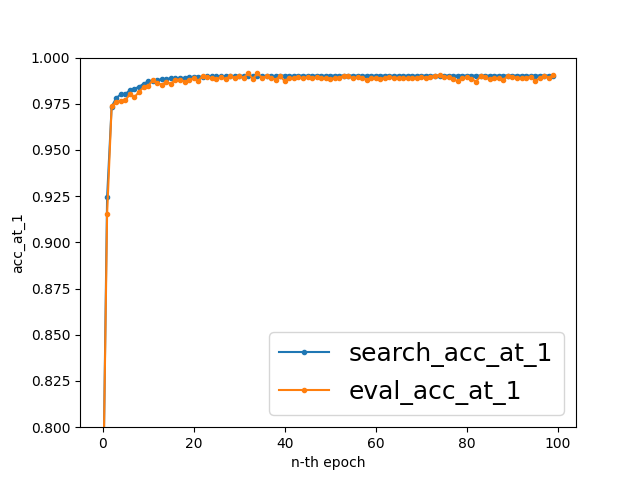}
		\caption{CWE-190}
	\end{subfigure}
	\begin{subfigure}[b]{0.48\linewidth}
		\includegraphics[width=\linewidth]{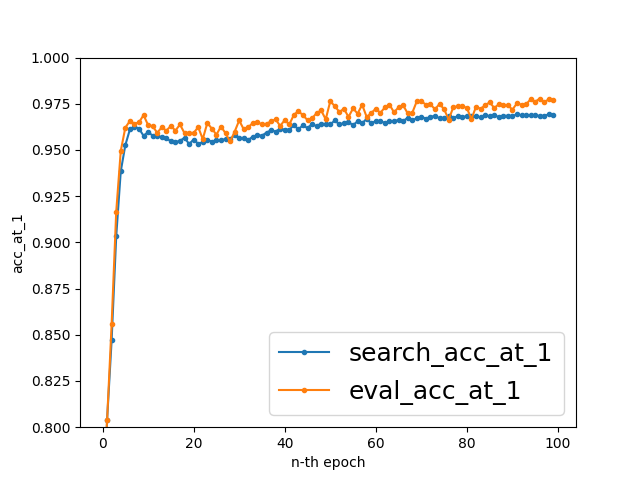}
		\caption{CWE-369}
	\end{subfigure}
	\begin{subfigure}[b]{0.48\linewidth}
		\includegraphics[width=\linewidth]{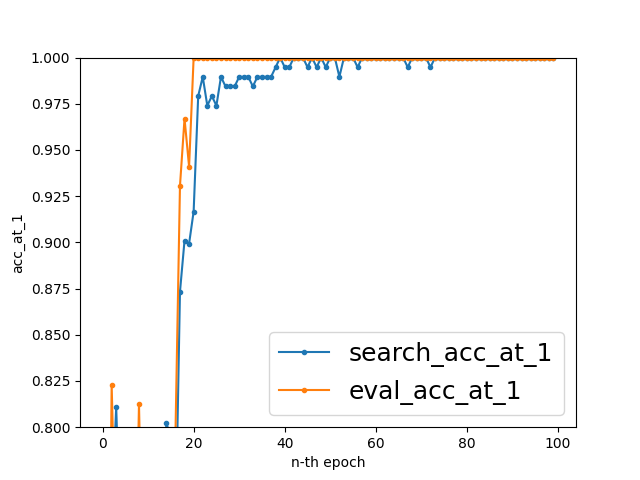}
		\caption{CWE-377}
	\end{subfigure}
	\caption{NAS-GDAS-JAE Cell Search Results - Part 1}
	\label{fig:NAS-result-1}
\end{figure}

\begin{figure}[]
	\centering
		\begin{subfigure}[b]{0.48\linewidth}
		\includegraphics[width=\linewidth]{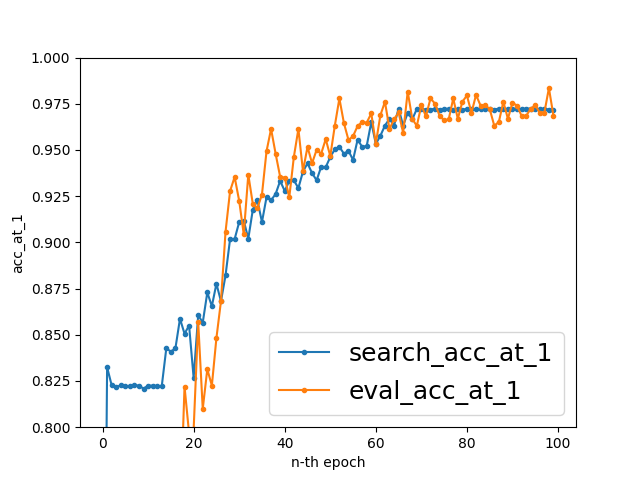}
		\caption{CWE-416}
	\end{subfigure}
	\begin{subfigure}[b]{0.48\linewidth}
		\includegraphics[width=\linewidth]{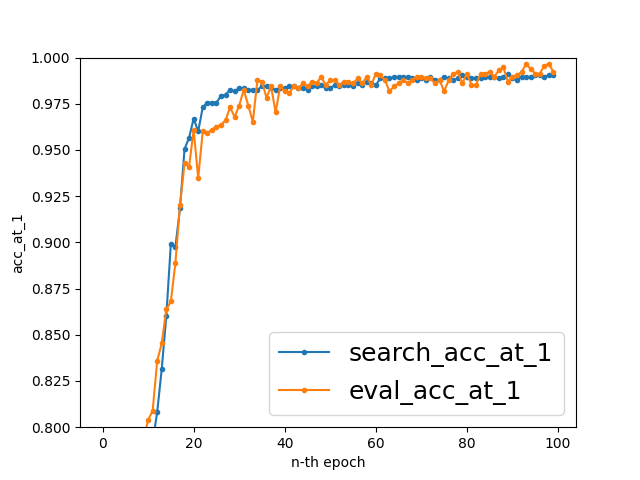}
		\caption{CWE-476}
	\end{subfigure}
	\begin{subfigure}[b]{0.48\linewidth}
	\includegraphics[width=\linewidth]{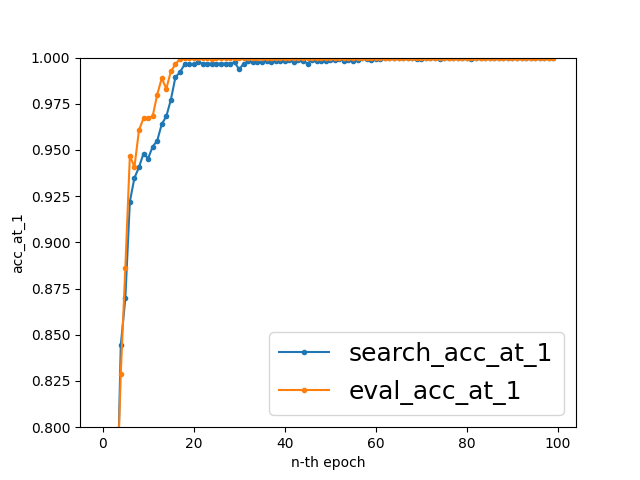}
	\caption{CWE-590}
\end{subfigure}
	\begin{subfigure}[b]{0.48\linewidth}
	\includegraphics[width=\linewidth]{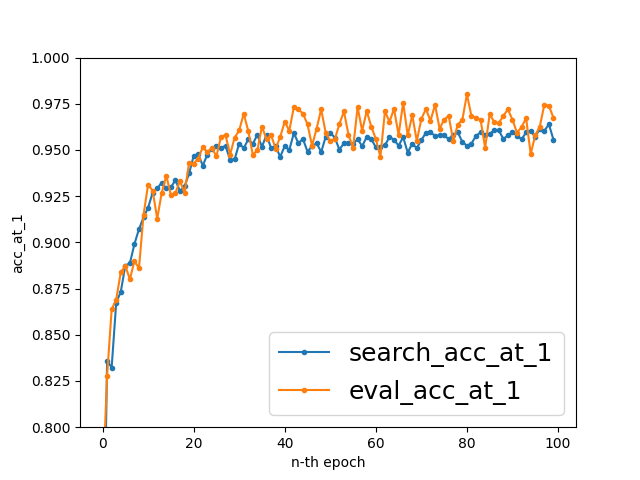}
	\caption{CWE-680}
\end{subfigure}
	\begin{subfigure}[b]{0.48\linewidth}
	\includegraphics[width=\linewidth]{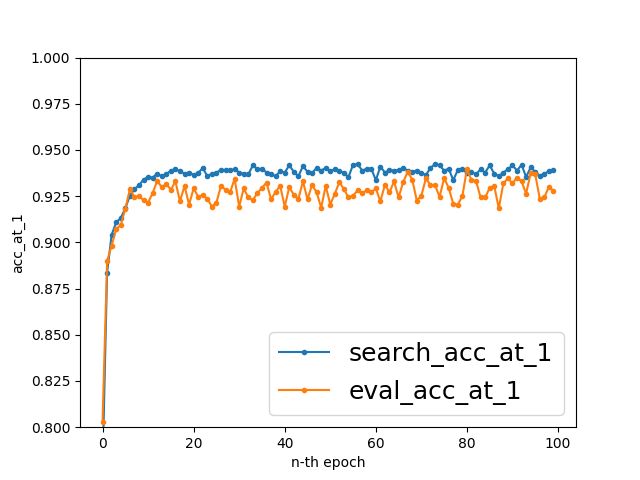}
	\caption{CWE-78}
\end{subfigure}
	\begin{subfigure}[b]{0.48\linewidth}
	\includegraphics[width=\linewidth]{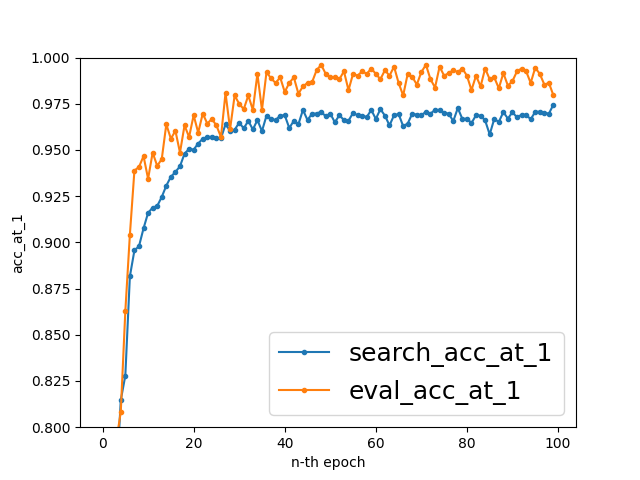}
	\caption{CWE-789}
\end{subfigure}
	\caption{NAS-GDAS-JAE Cell Search Results - Part 2}
	\label{fig:NAS-result-2}
\end{figure}